# Exploring Augmentation and Cognitive Strategies for AI based Synthetic Personae

Rafael Arias Gonzalez*, Simon Fraser University, Canada

Steve DiPaola, Simon Fraser University, Canada

**Abstract:** Large language models (LLMs) hold potential for innovative HCI research, including the creation of synthetic personae. However, their black-box nature and propensity for hallucinations pose challenges. To address these limitations, this position paper advocates for using LLMs as data augmentation systems rather than zero-shot generators. We further propose the development of robust cognitive and memory frameworks to guide LLM responses. Initial explorations suggest that data enrichment, episodic memory, and self-reflection techniques can improve the reliability of synthetic personae and open up new avenues for HCI research.

## 1 INTRODUCTION

Large Language Models (LLMs) present novel opportunities for Human-Computer Interaction (HCI) research. LLMs offer the potential for creating synthetic personae and generating synthetic data, potentially facilitating innovative explorations. However, their "black box" nature and tendency to produce hallucinations pose significant challenges for researchers.

While several techniques exist to reduce hallucination and increase explainability, these tend to come at other costs, such as model sizes or inference time.

In order to leverage LLMs as synthetic personae, this position paper argues for

- The use of LLMs as data augmenting systems, rather than zero-shot data generators, to maximize synthetic data generation.

- Designing more robust and efficient cognitive and memory frameworks for data retrieval and guided generation.

*Main Author

## 2 CHALLENGES IN LEVERAGING LLMS FOR HCI

### 2.1 Hallucination

Hallucination in LLMs occurs when models produce content that exhibits surface-level rationality and logic while being factually inaccurate or contradictory. The primary issue with hallucination is that these models produce inaccurate responses confidently, making it difficult to differentiate between trustworthy and false information. Hallucination as an inherent problem of LLMs has been widely documented [4, 6, 8].

The question of whether LLM hallucinations can be directly equated to human hallucinations remains open. Firstly, clinical definitions of human hallucination differ significantly from the phenomenon observed in LLMs [11]. While some researchers suggest alternative terminology like 'confabulation,' [11] we believe non-pathological terms like misattributions or false memories may be more analogous. Further investigation is required to better conceptualize LLM errors and to clarify the nature of LLM's inaccuracies and their potential relationship to human cognitive processes.

Various techniques exist to mitigate hallucination in dialogue generation. For knowledge-grounded dialogue (KGD), retrieval-augmented generation (RAG) methods have proven highly effective [10]. Here the model retrieves relevant knowledge before generating a response, helping reduce hallucinations while maintaining conversational coherence.

While simple RAG systems (which directly compare query and text embeddings) can lack precision and recall, newer RAG architectures offer significant improvements. These advanced models use techniques like chaining, re-ranking, or modularization [2] to deliver richer context for the LLM, but potentially increase processing time due to multiple LLM calls.

### 2.2 Memory and Explainability

Considering LLMs as synthetic personae within an HCI framework exposes a critical limitation: their lack of a persistent and grounded cognitive model. HCI research emphasizes the importance of understanding and modeling users' mental models, including their goals, beliefs, and decision-making processes. Without a robust internal representation of these elements, LLMs struggle to provide the level of consistency and explainability necessary for meaningful interaction in HCI contexts.

Traditional "guess the object" games provide a clear illustration of this challenge. Humans choose an object and store it in memory, ensuring consistency in their responses. Conversely, an LLM, which relies only on static weights and lacks persistent memory, may generate inconsistent answers that aren't linked to a specific object. This inconsistency highlights the absence of an internal cognitive model, preventing the LLM from maintaining a fixed target in line with how humans conceptualize the task.

This lack of persistent memory raises a concern regarding the authenticity of LLMs as synthetic personae. Even if an LLM's parameters enable some degree of internal reasoning, the explanations a model might offer for making specific decisions are generated on the fly when asked to articulate those processes post-generation. They were not explicitly encoded beforehand, given that there is no memory or update on the model's parameters. Consequently, an LLM's explanations might diverge from the actual reasoning encoded within its static parameters. These possible divergences suggest a potential disconnect between an LLM's expressed reasoning and the underlying computations driving its decisions.

Self-reflection mechanisms can partially address the issues of explainability and context-based reasoning (within the constraints of the model's window size) [3, 5, 7]. Models can be prompted to elucidate their internal processes or provide reasoning behind their outputs. This approach has demonstrated value in enhancing response quality. However, a notable trade-off exists: self-reflection can significantly increase computational overhead, given that the model must generate more information each time, slowing down the overall inference process

### 2.3 Real-world uses

Efforts to mitigate hallucination and enhance explainability in LLMs often come at the cost of increased inference times. This poses a distinct challenge when considering LLMs as synthetic personae, particularly in interactive contexts such as interviews or video game characters. In these scenarios, real-time responsiveness is crucial for maintaining a natural conversational flow or seamless gameplay experience. For example, a noticeable delay in response from a virtual therapist or an NPC (non-player character) could disrupt immersion and believability.

## 3  POTENTIAL STRATEGIES

### 3.1  LLMs for data augmentation

Recent research highlights the ability of Large Language Models (LLMs) to augment data for various NLP tasks. This includes generating conversational variations to improve model robustness [1], creating multilingual examples for better cross-lingual understanding [12], and rewriting examples for scarce-data scenarios to enhance specific word or phrase identification [13].

Given LLMs' robust data augmentation capabilities, their role as synthetic personae should be re-envisioned as augmenters rather than primary generators. Instead of expecting LLMs to generate inferences from minimal context (relying solely on internalized model training), providing them with substantial context for augmentation may better simulate the nuances of personae. In other words, we propose a paradigm in which we afford the model a defined structure to complete rather than expecting the model to generate complex content from scratch independently.

## 3.2 Cognitive and memory frameworks

To provide LLMs with richer context for character embodiment, we need frameworks that efficiently retrieve relevant data in an accessible format. Research on autonoetic consciousness, the ability to mentally re-experience past events, highlights the role of episodic memory and subjective experience in human conversation [9]. In contrast, traditional RAG systems lack this first-person perspective. To improve LLM performance, new memory frameworks should model information retrieval in a way that mirrors how humans dynamically access memories during interactions. Preemptively augmenting data with self-reflective content, such as diary entries or internal monologues, could provide RAG systems with readily accessible information rich in self-awareness, potentially enabling faster and more informed responses with a greater sense of self.

## 4 EXPLORATORY WORK

To explore the proposed solutions, we developed an episodic memory system integrated with a large language model (LLM). We selected the well-documented historical figure of Vincent Van Gogh as our test subject, leveraging the availability of his extensive biographical information. Our methodology consisted of the following phases:

### 4.1 Data Augmentation

To simulate autonoesis, we focused on enriching the source data with first-person perspectives and scene-specific context. We employed an LLM as a data augmentation tool, rewriting the entire biographical dataset to generate a movie script about Van Gogh. This script included a background summary, a narrator introduction, and first-person voiceovers of Van Gogh describing key life events. By providing the LLM with biographical data, we aimed to enhance its sense of self through the retrieved content.

We further augmented the biographical data using multiple LLM instances to extract and quantify relevant information from the generated script:

- Scene Analysis: An LLM, acting as a Van Gogh expert, analyzed each scene to identify key elements: characters present, dominant emotions, locations, and dates. Additionally, the expert provided a brief contextual summary, a relevance score, and a commentary for each scene.

- Emotional Quantification: We compiled a comprehensive list of emotions expressed throughout the script. A separate LLM instance assigned valence and arousal scores to each emotion, allowing us to calculate average valence and arousal scores for each scene.

- Standardization: LLMs were employed to reformat dates into a consistent format compatible with Python libraries for conversion into timestamps. Similarly, location descriptions were standardized to facilitate the extraction of latitude and longitude coordinates.

Our data augmentation process resulted in a comprehensive dataset. Each entry includes the following fields: scene background context, narrator introduction, Van Gogh's first-person narrative, general context, expert commentary, characters involved, valence and arousal scores, timestamp, latitude, longitude, and a relevance score. This representation provides rich contextual information for subsequent integration with the episodic memory system and LLM.

## 4.2  Episodic Memory Graph System

Our episodic memory model employs an adaptive graph-based structure. In order to obtain relevant retrieved data, we do the following:

1. Initial Query Matching: We leverage cosine similarity to identify candidate memory entries based on their contextual alignment with a given query. These entries serve as initial "entry points" into the graph.

2. Multi-Factor Ranking: To rank the remaining entries, we compound the cosine similarity with scores derived from other relevant metadata: Emotional Distance: We calculate the Euclidean distance between the entry point's emotional valence/arousal coordinates and each remaining entry. These distances are normalized (0-1), prioritizing entries with closer emotional profiles. Spatiotemporal Distance: Similarly, we normalize the distances in latitude/longitude and timestamps between the entry point and other entries.

3. Compound Score: We multiply the normalized values from each factor (emotional, spatiotemporal) with the initial cosine similarity score. This produces a final score that modifies the ranking, prioritizing entries not only based on semantic similarity but also on their emotional and spatiotemporal proximity to the query.

This multi-factor approach aims to refine retrieval beyond purely embedding similarity. By incorporating emotional and spatiotemporal context, we enable the system to surface memories more closely aligned with the affective and situational nature of the query.

## 4.3  LLM integration

Our LLM integration is streamlined and leverages the previously designed episodic memory system. It consists of three steps:

1. Memory Retrieval: Upon receiving a query, the episodic memory system retrieves relevant entries from the database, subject to two constraints: a maximum entry count and a similarity threshold. This ensures focused retrieval without compromising relevance.

Table 1. Sample answer excerpts to the query "Why did you cut your ear?"

| Model configuration | Response (excerpts) |
|---|---|
| Baseline LLM | (...) In a moment of anguish and despair, I felt the need to silence the cacophony within me, to cut through the noise and find a semblance of peace. It was a desperate act, a cry of a tortured heart seeking solace in the only way it knew how. (...) |
| Traditional RAG | Ah, the echo of that fateful night reverberates through the chambers of memory. (...) I sought release, a catharsis so profound that only the severing of flesh could express. (...) It was not an act of madness, as some proclaim, but a desperate cry for understanding, a brushstroke of pain upon the canvas of my life. cry for understanding, a brushstroke of pain upon the canvas of my life. |
| Augmented RAG (Autonoesis) | (...) It was a moment of anguish, of turmoil that I can hardly put into words. (...) It was a cry for help (...) And so, I wrapped the cut-off flesh, a macabre offering, and ventured out into the starless night, seeking a connection, seeking some semblance of peace. But alas, my dear friend Gauguin was not there (...) |
| Augmented RAG (Autonoesis + Ranked + Data) | (...) It was a moment of intense turmoil. I washed the cut-off flesh wrapped it carefully (...), I sought out Gauguin, my dear friend (...). But he was not there, and in my despair, I left my offering at the brothel. (...) Gauguin's absence, his significance in my life (...) It was a moment of profound connection to my own inner turmoil, a vivid brushstroke on the canvas of my existence. |

2. Context Construction: Retrieved entries are concatenated into a context prompt. We also include a general character description and task instructions (using standard prompt engineering techniques). To make the model adhere more to the augmented data, we also pass the raw retrieved values in the context (valence, arousal, relevance score). Finally, we include the chat history for as much as the context window allows. This provides the system with both short-term and long-term memory for each query.

3. Query Submission: The constructed context and query are then forwarded to the LLM for response generation.

## 5 PRELIMINARY FINDINGS

We conducted a comparative analysis of four system configurations to evaluate the impact of our approach:

- Baseline LLM: GPT-3.5 without any RAG integrations.
- Traditional RAG:: GPT-3.5 with RAG using the original biographical data.
- Augmented RAG (Autonoesis): GPT-3.5 with RAG using the LLM-generated autobiography (including scene context), simulating autonoesis.
- Augmented RAG (Autonoesis + Ranked + Data): GPT-3.5 with RAG using the LLM-generated autobiography, ranking entries, and incorporating the top entry's numerical data into the context.

Table 1 shows sample responses of the different systems from a query that we consider significant to Van Gogh's life: "Why did you cut your ear?".

### 5.1 Key Observations

Our analysis revealed that the baseline LLM offered poetic but incomplete responses, lacking narration. The traditional RAG system, while adhering to the narrative, lacked depth. From other experiments, we found it also exhibited inconsistent pronoun use, sometimes referring to the character in the third person. The simulated autonoesis RAG yielded richer responses, introducing contextually relevant characters (Gauguin). Lastly, combining autonoesis with ranking and numerical augmentation produced the most focused, informative, and explanatory responses. This demonstrates our approach's potential to provide rich context to the LLM, improving its ability to generate nuanced, accurate, and consistent responses within the Van Gogh persona.

## 6 DISCUSSION

Within the domain of HCI research, we argue that the most effective utilization of LLMs lies in their potential as data augmentation tools. Rather than relying on them for zero-shot generation, we propose the development of robust memory and cognitive frameworks to provide LLMs with richer context, going beyond the limitations of traditional RAG systems. Our experiments demonstrate that this augmentation and contextualization approach yields more informative and focused responses.

We envision several compelling applications of this approach. By augmenting real participant data, we can create synthetic personae with cognitive models that partially imitate the original participants. This opens the door to extensive interviews with these personae, even in scenarios that may be stressful or sensitive for human participants. Additionally, our system offers a degree of explainability by providing access to augmented data and ranked retrieved scenes. This transparency allows researchers to explore and understand the reasoning behind the model's responses, a crucial advantage in HCI research.

Our framework's emphasis on single RAG searches and ranking algorithms ensures fast response times, making it suitable for real-time interviews. Furthermore, by offloading some of the data processing and self-reflection from the model, we potentially allow for embedding smaller, more efficient models into systems where computational resources are constrained. This has particular relevance in industries such as video game development, where GPU limitations prohibit loading large models.

The findings discussed in this paper represent an initial exploration into the intricate relationship between data augmentation, cognitive modelling, and LLM performance. It highlights the promise of this field and underscores the need for further research. This work aims to spark new investigations, igniting a deeper understanding of how tailored data sets and advanced memory frameworks can unlock richer, more nuanced interactions with language models.